\documentclass{article}
\usepackage{spconf,amsmath,graphicx}
\usepackage{multirow}
\usepackage{booktabs}
\usepackage[table]{xcolor}
\usepackage{caption}
\usepackage{amsfonts}
\usepackage[numbers]{natbib}
\usepackage{subcaption}
\usepackage[flushmargin]{footmisc}
\usepackage[%
breaklinks=true,
colorlinks=true, 
citecolor=black,
urlcolor=black,
linkcolor=black,
linktocpage,
draft=false,
pageanchor=true,
hyperfigures=true,
plainpages=false
]{hyperref}

\title{CONFUSION2VEC 2.0: ENRICHING AMBIGUOUS SPOKEN LANGUAGE REPRESENTATIONS WITH SUBWORDS}
\name{Prashanth Gurunath Shivakumar\thanks{\texttt{email: pgurunat@usc.edu}}, Panayiotis Georgiou, Shrikanth Narayanan}
\address{University of Southern California, Los Angeles, California, USA}
\begin{document}
%
\maketitle
\begin{abstract}
Word vector representations enable machines to encode human language for spoken language understanding and processing.
Confusion2vec, motivated from human speech production and perception, is a word vector representation which encodes ambiguities present in human spoken language in addition to semantics and syntactic information.
Confusion2vec provides a robust spoken language representation by considering inherent human language ambiguities.
In this paper, we propose a novel word vector space estimation by unsupervised learning on lattices output by an automatic speech recognition (ASR) system.
We encode each word in confusion2vec vector space by its constituent subword character n-grams.
We show the subword encoding helps better represent the acoustic perceptual ambiguities in human spoken language via information modeled on lattice structured ASR output.
The usefulness of the proposed Confusion2vec representation is evaluated using semantic, syntactic and acoustic analogy and word similarity tasks.
We also show the benefits of subword modeling for acoustic ambiguity representation on the task of spoken language intent detection.
The results significantly outperform existing word vector representations when evaluated on erroneous ASR outputs.
We demonstrate that Confusion2vec subword modeling eliminates the need for retraining/adapting the natural language understanding models on ASR transcripts.
\end{abstract}
\section{Introduction}
\label{sec:intro}
Speech is the primary and most natural mode of communication for humans.
This makes its use also attractive for human computer interaction, which in turn requires decoding human language to enable spoken language understanding.
Human language is a complex construct involving multiple dimensions of information including semantics, syntax and often contain ambiguities which make it challenging for machine inference of communication intent, emotions etc.
Several word vector representations have been proposed for effectively describing the human language in the natural language processing community.

Contextual modeling techniques like language modeling, i.e., predicting the next word in the sentence given a window of preceding context, have been shown to model meaningful word representations \cite{bengio2003neural,mikolov2010recurrent}.
Bag-of-word based contextual modeling, where the current word is predicted given both its left and right (local) contexts has shown to capture language semantics and syntax \cite{mikolov2013efficient}.
Similarly, predicting local context from the current word, referred to as skip-gram modeling, is shown to better represent semantic and syntactic distances between words \cite{mikolov2013distributed}.
In \cite{pennington2014glove} log bi-linear models combining global word co-occurrence information and local context information, termed as global vectors (GloVe), is shown to produce meaningful structured vector space.
Bi-directional language models are proposed in \cite{peters2018deep}, where internal states of deep neural networks are combined to model complex characteristics of word use and its variance over linguistic contexts.
The advantages of bi-directional modeling are further exploited along with self-attention using transformer networks \cite{vaswani2017attention} to estimate a representation, termed as BERT (Bidirectional Encoder Representations from Transformers), that has shown its utility on a multitude of natural language understanding tasks \cite{devlin2018bert}.
Models such as BERT, ELMo estimate word representations that vary depending on the context, whereas the context-free representations including GloVe and Word2Vec generate a single representation irrespective of the context.

However, most of the word vector representations infer the knowledge through contextual modeling and many of the inherent ambiguities present in human language are often unrecognized or ignored.
For instance, from the perspective of spoken language, the ambiguities can be associated with how similar the words sound, i.e., for example, the words ``see'' and ``sea'' sound acoustically identical but have different meanings.
The ambiguities can also be associated with the underlying speech signal itself due to wide range of acoustic environments involving noise, overlapped speech and channel, room characteristics.
These ambiguities often project themselves as errors through ASR systems.
Most of the existing word vector representations such as word2vec \cite{mikolov2013distributed,mikolov2013efficient}, fasttext \cite{bojanowski2017enriching}, GloVe \cite{pennington2014glove}, BERT \cite{devlin2018bert}, ELMo \cite{peters2018deep} do not account for the ambiguities present in speech signals and thus degrade while processing the output of noisy ASR transcripts.

Confusion2vec was recently proposed to handle representation ambiguity information present in human language \cite{shivakumar2019confusion2vec}.
Confusion2vec is estimated by unsupervised skip-gram training on the ASR output lattices and confusion networks.
The analysis of inherent acoustic ambiguity information of the embeddings displayed meaningful interactions between the semantic-syntactic subspace and acoustic similarity subspaces.
In \cite{shivakumar2019spoken}, the usefulness of the Confusion2vec was confirmed on the task of spoken language intent detection.
The Confusion2vec representation significantly outperformed typical word embeddings including word2vec and GloVe when evaluated on noisy ASR transcripts by reducing the classification error rate by approximately 20\% relative.

Although, there have been few attempts in leveraging information present in word lattices and word confusion networks for several tasks \cite{tai2015improved,ladhak2016latticernn,tan2018lattice,xiao2019lattice,sperber2019self,huang2020learning}, the main downside with these works is that the word representation estimated by such techniques are task dependent and are restricted to a particular domain and dataset.
Moreover, availability of most of the task specific datasets are limited and task specific speech data are expensive to collect.
The advantage of Confusion2Vec is that it estimates a generic, task-independent word vector representation via unsupervised learning on lattices or confusion networks generated by an ASR on any speech conversations.

In this paper, we incorporate subwords to represent each word for modeling both the acoustic ambiguity information and the contextual information.
Each word is modeled as a sum of constituent n-gram characters.
Our motivation behind the use of subwords are the following: (i) they incorporates morphological information of the words by encoding internal structure of words \cite{bojanowski2017enriching}, (ii) the bag of character n-grams often have a high overlap between acoustically ambiguous words, (iii) subwords help model under-represented words more efficiently, thereby leading to more robust estimation with limited available data, which is the case since training Confusion2Vec is restricted to ASR lattice outputs, (iv) subwords enable representations for out-of-vocabulary words which are common-place with end-to-end ASR systems outputting characters.

The rest of the paper is organized as follows: Confusion2vec is introduced in Section~\ref{sec:c2v}.
The proposed subword modeling is presented in Section~\ref{sec:c2v_subword}.
Section~\ref{sec:eval} gives details of the evaluation techniques employed for assessing the word embedding models.
The experimental setup and results of various analogy and similarity tasks are presented in section~\ref{sec:result_1}.
Section~\ref{sec:slu} presents the application of the proposed word vector representation to the spoken language intent detection task. 
Finally, the paper is concluded in section~\ref{sec:conclusion}.

\begin{figure*}[t]
\centering
\includegraphics[width=0.6\textwidth]{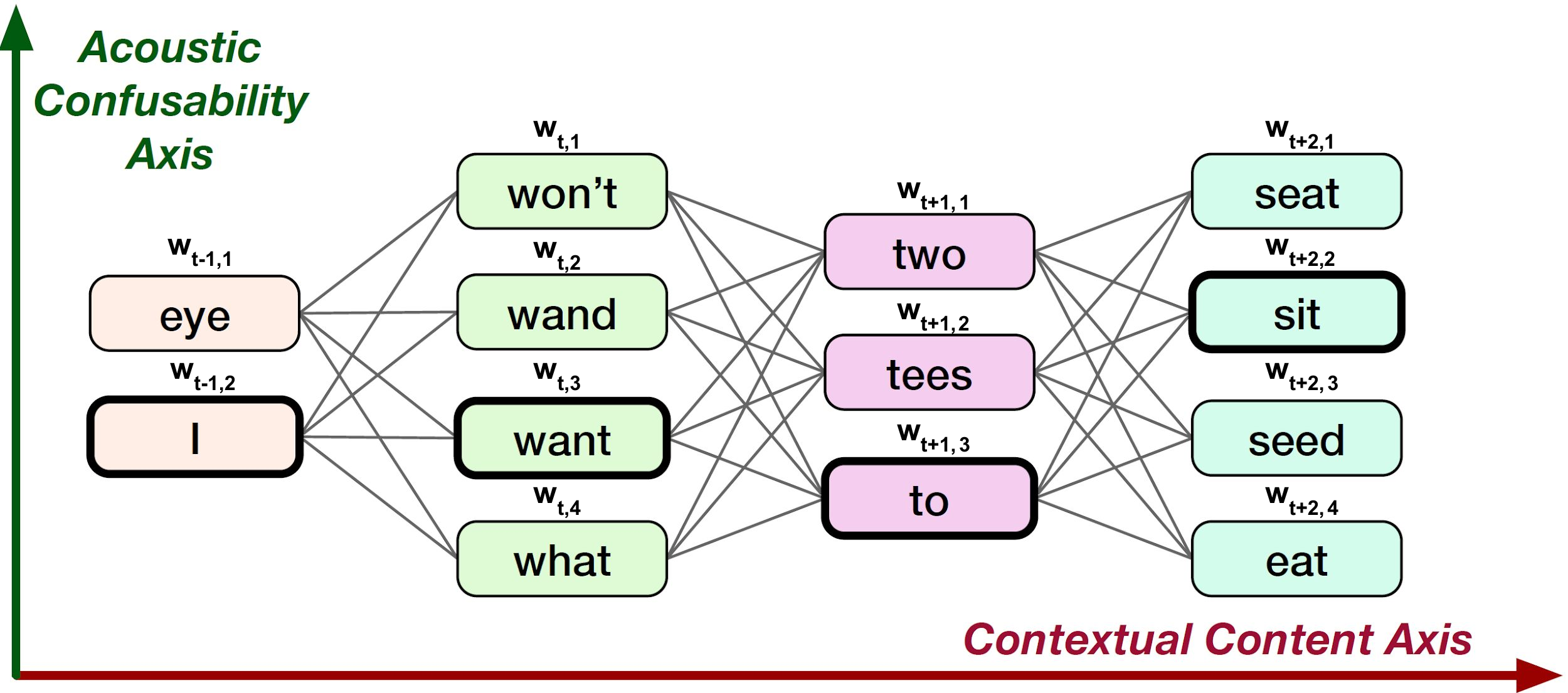}
\captionsetup{justification=centering}
\caption{Example Confusion Network Output by ASR for the ground-truth phrase ``I want to sit''}\label{fig:confusion_network}
\end{figure*}

\section{Confusion2Vec}\label{sec:c2v}
In psycho-acoustics, it is established that humans also relate words with how they sound \cite{aydelott2004effects} in addition to semantics and syntax.
Inspired by principles of human speech production and perception, we previously proposed Confusion2vec \cite{shivakumar2019confusion2vec}.
The core idea is to estimate a hyper-space that not only captures the semantics and syntax of human language, but also augments the vector space with acoustic ambiguity information, i.e., word acoustic similarity information.
In other words, word2vec, GloVe can be viewed as a subspace of the confusion2vec vector space.

Several different methodologies are proposed for capturing the ambiguity information.
The methodologies are an adaptation of the skip-gram modeling for word confusion networks or lattice-like structures.
The word lattices are directed acyclic weighted graphs of all the word sequences that are likely possible.
A confusion network is a specific type of lattice with constraints that each word sequence passes through each node of graph.
Such lattice-like structures can be derived from machine learning algorithms that output probability measures, for example, an ASR.
Figure~\ref{fig:confusion_network}, illustrates a confusion network that can possibly result from a speech recognition system.
Unlike typical simple sentences which are used for training word embeddings like word2vec, GloVe, BERT, ELMo etc., the information in the confusion network can be viewed along two dimensions: (i) contextual dimension, and (ii) acoustic ambiguity dimension.

More specifically, four different configurations of skip-gram modeling algorithms are proposed in our previous work \cite{shivakumar2019confusion2vec}, namely: (i) top-confusion, (ii) intra-confusion, (iii) inter-confusion, and (iv) hybrid model.
The top-confusion version considers only the most-probable path of the ASR confusion network and applies the typical skip-gram model on it.
The intra-confusion version applies the skip-gram modeling on the acoustic ambiguity dimension of the confusion network and ignores the contextual information, i.e., each ambiguous word alternative is predicted by the other over a pre-defined local context.
The inter-confusion version applies the skip-gram modeling on the contextual dimension but over each of the acoustic ambiguous words.
The hybrid model is a combination of both the intra and inter-confusion configurations.
More information on the training configuration is available in \cite{shivakumar2019confusion2vec}.
The present work builds upon this basic Confusion2vec framework.

\section{Confusion2Vec 2.0 subword model}\label{sec:c2v_subword}
Subword encoding of words has been popular in modeling semantics and syntax of language using word vector representations \cite{bojanowski2017enriching,devlin2018bert,peters2018deep}.
The use of subwords are mainly motivated by the fact that the subwords incorporate morphological information which can be helpful, for example, in relating the prefixes, suffixes and the word root.
In this work, we apply subword representation for encoding the word ambiguity information in the human language.
We believe we have a compelling case for the use of subwords for representing the acoustic similarities (ambiguities) between the words in the language since more similarly sounding words often have highly overlapping subword representations.
This helps model the level of overlap and estimate the magnitude of acoustic similarity robustly.
Moreover, use of subwords should help in efficient encoding of under-represented words in the language. 
This is crucial in the case of Confusion2vec because we are restricted to speech data and their corresponding decoded ASR lattices for training, thereby limiting word-word co-occurrence in contrast to typical word vector representation which can be trained on large amounts of easily available plain text data.
Another important aspect is the ability to represent out-of-vocabulary words which are a common place occurrence with end-to-end ASR systems outputting character sequences.

In the proposed model, each word $w$ is represented as a sum of its constituent n-gram character subwords.
This enables the model to infer the internal structure of each word.
For example, a word ``\texttt{want}`` is represented with the vector sum of the following subwords:
\begin{center}
\texttt{$<$wa, wan, ant, nt$>$, $<$wan, want, ant$>$, $<$want, want$>$, $<$want$>$}
\end{center}
Symbols $<$ and $>$ are used to represent the beginning and end of the word.
The n-grams are generated for n=3 upto n=6.
It is apparent that an acoustically ambiguous, similar sounding word ``\texttt{wand}'' has a high degree of overlap with the set of n-gram characters.

In this paper, we consider two modeling variations: (i) inter-confusion, and (ii) intra-confusion versions of confusion2vec with the subword encoding.

\subsection{Intra-Confusion Model}
The goal of the intra-confusion model is to estimate the inter-word relations between the acoustically ambiguous words that appear in the ASR lattices.
For this, we perform skip-gram modeling over the acoustic similarity dimension (see Figure~\ref{fig:confusion_network}) and ignore the contextual dimension of the utterance.
The objective of the intra-confusion model is to maximize the following log-likelihood:
\begin{equation}\label{eq:c2v-a}
\sum_{t=1}^{T}\sum_{\hat{a} \in \hat{A_t}}\sum_{a\in A_t} log~p(w_{t,a} | w_{t,\hat{a}})
\end{equation}
where $T$ is the length of the utterance (confusion network) in terms of number of words, $w_{i,j}$ is the word in the confusion network output by the ASR at time-step $i$ and $j$ is the index of the word among the ambiguous alternatives.
$\hat{A_t}$ is the set of indices of all ambiguous words at time-step $t$, $\hat{a}$ is the index of the current word along the acoustic ambiguity dimension, $A_t\subseteq\hat{A_t}-\hat{a}$ is the subset of ambiguous words barring $\hat{a}$ at the current word $t$, i.e., for example from Figure~\ref{fig:confusion_network}, for the current word, $w_{t,\hat{a}}$, ``\texttt{want}'', $A_t\subseteq$ {\small\{\texttt{wand, won't, what}\}}.
Additionally, for subword encoding, each word input is represented as:
\begin{equation}\label{eq:subword}
w_{i,j} = \sum_{s \in S_w}x_s
\end{equation}
where $S_w$ is the set of all character n-grams ranging from n=3 to n=6 and the word itself and $x_s$ is the vector representation for n-gram subword $s$.
Few training samples (input, target) generated for this configuration pertaining to input confusion network in Figure~\ref{fig:confusion_network} are {\small\texttt{(I, eye), (eye, I), (want, wand), (want, won't), (won't, what), (wand, what)}} etc.

\subsection{Inter-Confusion Model}
The aim of the inter-confusion model is to jointly model the contextual co-occurrence information and the acoustic ambiguity co-occurrence information along both the axis depicted in the confusion network.
Here, the skip-gram modeling is performed over time context and over all the possible acoustic ambiguities.
The objective of the inter-confusion model is to maximize the following log-likelihood:
\begin{equation}\label{eq:c2v-c}
\sum_{t=1}^T\sum_{\hat{a} \in \hat{A_t}}\sum_{c\in C_t}\sum_{a\in A_c} log~p(w_{c,a}|w_{t,\hat{a}})
\end{equation}
where $C_t$ corresponds to set of indices of nodes of confusion network, i.e., words around the current word $t$ along the time-axis and $c$ is the current context index.
$A_c$ is the set of indices of acoustically ambiguous words at a context $c$.
For example, for the current word, $w_{t,\hat{a}}$, ``\texttt{want}'' in Figure~\ref{fig:confusion_network}, $A_c\subseteq$ {\small\{\texttt{I, eye, two, tees, to, seat, sit, seed, eat}\}} and $A_t\subseteq$ {\small\{\texttt{wand, won't, what, want}\}}.
Note, each word input is subword encoded as in equation~\ref{eq:subword}.
Few training samples (input, target) generated for this configuration are {\small\texttt{(want, I), (want, eye), (want, two), (want, to), (want, tees), (what, I), (what, eye), (what, to), (what, tees), (what, two), (won't, eye)}} etc.

\subsection{Training Loss and Objective}
Negative sampling is employed for training the embedding model.
Negative sampling was first introduced for training word2vec representation \cite{mikolov2013distributed}.
It is a simplification of the Noise Contrastive Estimation objective \cite{gutmann2012noise}.
The negative sampling for training the embedding can be posed as a set of binary classification problems which operates on two classes: presence of signal or absence (noise).
In the context of word embeddings the presence of the context words are treated as positive class and the negative class is randomly sampled from the unigram distribution of the vocabulary.
The negative sampling for subword model can be expressed using binary logistic loss as:
\begin{equation}
log~\sigma(\sum_{s \in S_{w_i}}x_{s}^To_{w_t}) + \sum_{k=1}^K \mathbb{E}_{w_{k}\sim P_n (w)} log~\sigma(-\sum_{s \in S_{w_i}}x_s^To_{w_k})
\end{equation}
where $\sigma(x) = \frac{1}{1+e^{-x}}$, $w_i$ is the input word, $w_t$ is the output word, $S_{w_{i}}$ is the set of n-gram character subwords for the word $w_i$, $x_{s}$ is the vector representation for the character n-gram subword $s$ and $o_{w_t}$ is the output vector representation of target word $w_t$. 
$K$ is the number of negative samples to be drawn from the negative sample, noise distribution $P_n (w)$.
The noise distribution $P_n (w)$ is chosen to be the unigram distribution of words in the vocabulary raised to the $3/4^{\text{th}}$ power as suggested in \cite{mikolov2013distributed}.
Note, for confusion2vec the input word $w_i$ and target word $w_t$ are derived according to equations~\ref{eq:c2v-a} and ~\ref{eq:c2v-c} for implementing the respective training configurations 

\section{Evaluations}\label{sec:eval}
We perform evaluations of the proposed word embeddings along two aspects.
One, in view of the assessing the useful, meaningful information embedded in the word vector representation.
Second, in its application to a realistic task of spoken language intent detection.

\subsection{Analogy and Similarity Tasks}
For evaluating the inherent semantic and syntactic knowledge of the word embeddings, we employ two tasks: (i) semantic-syntactic analogy task, and (ii) word similarity task.
The word analogy task was first proposed in \cite{mikolov2013efficient} which comprises word pair analogy questions of the form $W_1$ is to $W_2$ as $W_3$ is to $W_4$.
The analogy is answered correct if $vec(W_1) - vec(W_2) + vec(W_3)$ is most similar to $vec(W_4)$.
Another prominent approach is the word similarity task, where rank-correlation between cosine similarity of set of pair of word vectors and human annotated word similarity scores are assessed \cite{schnabel2015evaluation}.
For word similarity task, we use the WordSim-353 database \cite{finkelstein2001placing} consisting of 353 pairs of words annotated over a score of 1 to 10 depending on the magnitude of word similarity as perceived by humans.

For assessing the word acoustic ambiguity (similarity) information, we conduct the acoustic analogy task, Semantic\&syntactic–acoustic analogy task and Acoustic similarity tasks proposed in \cite{shivakumar2019confusion2vec}.
The Acoustic analogy task comprises of word pair analogies compiled using homophones which answer questions of the form: $W_1$ sounds similar to $W_2$ as $W_3$ sounds similar to $W_4$.
The acoustic analogy task is designed to assess the ambiguity information embedded in the word vector space \cite{shivakumar2019confusion2vec}.
The semantic\&syntactic-acoustic analogy task is designed to assess semantic, syntactic and acoustic ambiguity information simultaneously.
The analogies are formed by replacing certain words by their homophone alternatives in the original semantic and syntactic analogy task \cite{shivakumar2019confusion2vec}.
The acoustic word similarity task is analogous to the word similarity task, i.e., it contains of word pairs which are rated on their acoustic similarity based on the normalized phone edit distances.
A value of 1.0 refers to two words sounding identical and 0.0 refers to the word pairs being acoustically dissimilar.
More details regarding the evaluation methodologies are available in \cite{shivakumar2019confusion2vec}.
The evaluation datasets are made available\footnote{\url{https://github.com/pgurunath/confusion2vec_2.0}}.

\begin{table*}[t]
\begin{center}
\resizebox{\textwidth}{!}{%
\begin{tabular}{llcccccc} 
\toprule
& \multirow{2}{*}{Model} & \multicolumn{4}{c}{Analogy Tasks} & \multicolumn{2}{c}{Similarity Tasks} \\
\cmidrule(lr){3-6} \cmidrule(lr){7-8}
& & S\&S & Acoustic & S\&S-Acoustic & Average Accuracy & Word Similarity & Acoustic Similarity \\
\midrule
 & Google W2V \cite{mikolov2013distributed} & 61.42\% & 0.9\% & 16.99\% & 26.44\% & \textbf{0.6893} & -0.3489 \\
& In-domain W2V & 59.17\% & 0.6\% & 8.15\% & 22.64\% & 0.4417 & -0.4377 \\
& fastText \cite{bojanowski2017enriching} & \textbf{75.93\%} & 0.46\% & 17.40\% & 31.26\% & 0.7361 & -0.3659 \\
\midrule
\multirow{2}{*}{\shortstack[l]{\textbf{Confusion2Vec 1.0}\\(word) \cite{shivakumar2019confusion2vec}}} & C2V-a & 63.97\% & 16.92\% & 43.34\% & 41.41\% & 0.5228 & 0.6200 \\
& C2V-c & \textbf{65.45\%} & 27.33\% & 38.29\% & 43.69\% & 0.5798 & 0.5825 \\
\midrule
\textbf{Confusion2Vec 2.0} & C2V-a & 56.74\% & \textbf{50.79\%} & \textbf{44.67\%} & \textbf{50.73\%} & 0.3181 & \textbf{0.8108} \\
(subword) & C2V-c & 56.87\% & \textbf{51.00\%} & \textbf{44.98\%} & \textbf{50.95\%} & 0.2893 & \textbf{0.8106} \\
\bottomrule
\end{tabular}}
\end{center}
\captionsetup{justification=centering}
\caption{\textbf{Results: Different proposed models}\\
\small{\textbf{C2V-a: Intra-Confusion, C2V-c: Inter-Confusion, S\&S: Semantic \& Syntactic Analogy.}\\
For the analogy tasks: the accuracies of baseline word2vec models are for top-1 evaluations, whereas of the other models are for top-2 evaluations (as discussed in \cite{shivakumar2019confusion2vec}).
For the similarity tasks: all the correlations (Spearman's) are statistically significant with $p<0.001$.}}\label{tab:results}
\end{table*}

\subsection{Spoken Language Intent Classification}
We also evaluate the efficacy of the proposed word representation models on the task of spoken language intent classification.
A recurrent neural network (RNN) based classifier is employed by initializing the embedding layer with the proposed word vectors.
Classification experiments are conducted by training the recurrent neural network on (i) clean manual transcripts, and (ii) noisy ASR transcripts, with evaluations on both manual and ASR transcripts.
Classification error rates of the intent detection is used to derive assessments of the word vector representations.

\section{Analogy \& Similarity Tasks}\label{sec:result_1}
\subsection{Database}
The Fisher English Training Part 1, Speech (LDC2004S13) and Fisher English Training Part 2, Speech (LDC2005S13) corpora \cite{cieri2004fisher} are used for both training the ASR and the confusion2vec 2.0 embeddings.
The choice of database is based on \cite{shivakumar2019confusion2vec} for direct comparison purposes.
The corpus consists of spontaneous telephonic conversations between 11,972 native English speakers.
The speech data amounts to approximately 1,915 hours sampled at 8 kHz.
The corpus is divided into 3 parts for training (1,905 hours, 1,871,731 utterances), development (5 hours, 5000 utterances) and test (5 hours, 5000 utterances).
Overall, the transcripts contain approximately 20.8 million word tokens and vocabulary size of 42,150.

\subsection{Experimental Setup}
The experimental setup is maintained identical to \cite{shivakumar2019confusion2vec} for direct comparison.
Brief detail of the setup is as follows:

\subsubsection{Automatic speech recognition}\label{sec:asr}
A hybrid HMM-DNN based acoustic model is trained on the train subset of the speech corpus using the KALDI speech recognition toolkit \cite{povey2011kaldi}.
40 dimensional mel frequency cepstral coefficients (MFCC) features are extracted along with the i-vector features for training the acoustic model.
The i-vector features are used to provide speaker and channel characteristics to aid acoustic modeling.
The DNN acoustic model, comprises 7 layers with P-norm non-linearity (p=2) each with 350 units \cite{zhang2014improving}.
The DNN is trained using 5 MFCC frame splices with left and right context of 2 to classify among 7979 Gaussian mixtures with stochastic gradient descent optimizer.
The CMU pronunciation dictionary \cite{weide1998cmu} is used as the word-pronunciation transcription lexicon.
A tri-gram language model is trained on the training subset of the Fisher English Speech Corpus.
The ASR yields word error rates (WER) of 16.57\% and 18.12\% on the development and the test datasets.
Lattices are derived during the ASR decoding with a decoding beam size of 11 and lattice beam size of 6.
The lattices are converted to confusion networks with the minimum Bayes risk criterion \cite{xu2011minimum} for training the confusion2vec embeddings.
The resulting confusion networks have a vocabulary size of 41,274 and 69.5 million words, with an average of 3.34 alternative (ambiguous) words for each edge in the graph.

\subsubsection{Confusion2Vec 2.0} 
In order to train the embedding, most frequent words are sub-sampled as suggested in \cite{mikolov2013distributed}, with the rejection threshold set to $10^{-4}$.
Also, a minimum frequency threshold of 5 is set and the rarely occurring words are pruned from the vocabulary.
The context window size for both the acoustic ambiguity and contextual dimensions are uniformly sampled between 1 and 5.
The dimension of the word vectors are set to 300.
The number of negative samples for negative sampling is chosen to be 64.
The learning rate is set to 0.01 and trained for a total of 15 epochs using stochastic gradient descent.
All the hyper-parameters are empirically chosen for optimal performance on the development set.
We implemented the confusion2vec 2.0 by modifying the source code from fastText\footnote{\texttt{https://github.com/facebookresearch/fastText}} \cite{bojanowski2017enriching}.
We make our source code and trained models available at \url{https://github.com/pgurunath/confusion2vec_2.0}.

\begin{table*}[t]
\begin{center}
\resizebox{\textwidth}{!}{%
\begin{tabular}{llcccccc} 
\toprule
& \multirow{2}{*}{Model} & \multicolumn{4}{c}{Analogy Tasks} & \multicolumn{2}{c}{Similarity Tasks} \\
\cmidrule(lr){3-6} \cmidrule(lr){7-8}
& & S\&S & Acoustic & S\&S-Acoustic & Average Accuracy & Word Similarity & Acoustic Similarity \\
\midrule
 & Google W2V \cite{mikolov2013distributed} & 61.42\% & 0.9\% & 16.99\% & 26.44\% & \textbf{0.6893} & -0.3489 \\
& In-domain W2V & 59.17\% & 0.6\% & 8.15\% & 22.64\% & 0.4417 & -0.4377 \\
& fastText \cite{bojanowski2017enriching} & 75.93\% & 0.46\% & 17.40\% & 31.26\% & 0.7361 & -0.3659 \\
\midrule
\multirow{2}{*}{\shortstack[l]{\textbf{Confusion2Vec 1.0}\\(word) \cite{shivakumar2019confusion2vec}}} & C2V-1 + C2V-a & 67.03\% & 25.43\% & 40.36\% & 44.27\% & 0.5102 & 0.7231 \\
& C2V-1 + C2V-c & 70.84\% & 35.25\% & 35.18\% & 47.09\% & 0.5609 & 0.6345 \\
\cmidrule(lr){2-8}
& C2V-1 + C2V-c (JT) & 65.88\% & \textbf{49.4\%} & 41.51\% & \textbf{52.26\%} & 0.5379 & \textbf{0.7717} \\
\midrule
\textbf{Confusion2Vec 2.0} & fastText + C2V-a & \textbf{76.10\%} & 22.67\% & \textbf{49.15\%} & \textbf{49.31\%} & 0.5744 & \textbf{0.7577} \\
(subword) & fastText + C2V-c & \textbf{76.16\%} & 22.56\% & \textbf{49.12\%} & \textbf{49.12\%} & 0.5732 & \textbf{0.7573} \\
\bottomrule
\end{tabular}}
\end{center}
\captionsetup{justification=centering}
\caption{\textbf{Results: Different proposed models}\\
\small{\textbf{C2V-a: Intra-Confusion, C2V-c: Inter-Confusion, S\&S: Semantic \& Syntactic Analogy.}\\
For the analogy tasks: the accuracies of baseline word2vec models are for top-1 evaluations, whereas of the other models are for top-2 evaluations (as discussed in \cite{shivakumar2019confusion2vec}).
For the similarity tasks: all the correlations (Spearman's) are statistically significant with $p<0.001$.}}\label{tab:concat_results}
\end{table*}

\subsection{Results}\label{sec:results_1}
Table~\ref{tab:results} lists the results in terms of accuracies for analogy tasks and rank-correlations for similarity tasks.
The first two rows correspond to results with the original word2vec.
Google W2V model is the open source model released by Google\footnote{\texttt{https://code.google.com/archive/p/word2vec/}}, trained on 100 billion word Google News dataset.
We also train an in-domain version of original word2vec on the Fisher English corpus for fair comparison with the confusion2vec models, referred to as ``In-domain W2V'' in Table~\ref{tab:results}.
The fastText model employed is the open source model trained on Wikipedia dumps with a vocabulary size of more than 2.5 million words released by Facebook\footnote{\texttt{https://fasttext.cc/docs/en/pretrained-vectors.html}}.
The middle two rows of the table correspond to confusion2vec embeddings without subword encoding and they are taken directly from \cite{shivakumar2019confusion2vec}.
The bottom two rows correspond to the results obtained with subword encoding.
Note, the confusion2vec 1.0 is initialized on the Google word2vec model for better convergence.
The confusion2vec 2.0 model is initialized on the fastText model to maintain compatibility with subword encodings.
We normalize the vocabulary for all the experiments, meaning the same vocabulary is used to evaluate the analogy and similarity tasks to allow for fair comparisons.

Comparing the baseline word2vec and fastText embeddings to the confusion2vec, we observe the baseline embeddings perform well on the semantic\&syntactic analogy task and provide good positive correlation on the word similarity task as expected.
However, they perform poorly on the acoustic analogy task, semantic\&syntactic-acoustic analogy task and give small negative correlation on the acoustic analogy task.
All the confusion2vec models perform relatively well on the semantic\&syntactic analogy task and word similarity task, but more importantly, yield high accuracies on acoustic analogy task and semantic\&syntactic-acoustic analogy tasks and provide high positive correlation with the acoustic similarity task.

Specifically with Confusion2Vec 2.0, among the analogy tasks, we observe the subword encoding enhances the acoustic ambiguity modeling.
For the acoustic analogy task we find relative improvement of upto 46.41\% over its non-subword counterpart.
Moreover, even for the semantic\&syntactic-acoustic analogy task, we observe improvements with subword encoding.
However, we find a small reduction in performance for the original semantic and syntactic analogy task.
Regardless of the small dip in the performance, the accuracies remain acceptable in comparison to the in-domain word2vec model.
Overall, taking the average accuracy of all the analogy tasks, we obtain an increase of approximately 16.62\% relative over the non-subword confusion2vec models.

Investigating the results for the similarity tasks, we find significant and high correlation of 0.81 for acoustic similarity task with the subword encoding.
Again, a small degradation is observed with the word similarity task obtaining a correlation of 0.3181 against the 0.4417 of the in-domain baseline word2vec model.
Overall, the results of the analogy and the similarity tasks suggest the subword encoding greatly enhances the ambiguity modeling of confusion2vec.

\begin{figure*}[t]
\centering
\begin{subfigure}{.49\textwidth}
\centering
\includegraphics[width=0.95\textwidth]{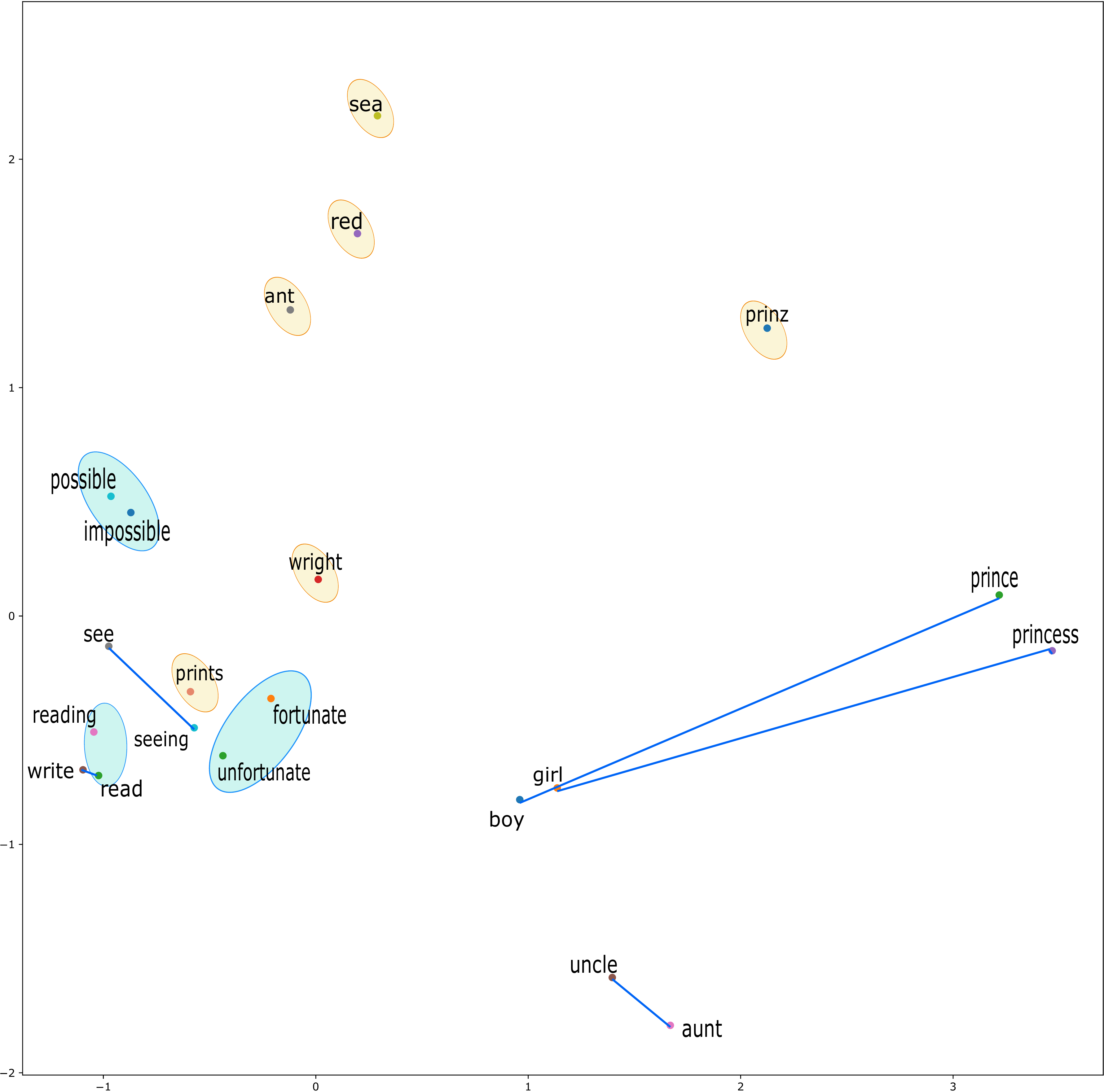}
\caption{fastText}\label{fig:pca_fasttext}
\end{subfigure}
\begin{subfigure}{.49\textwidth}
\centering
\includegraphics[width=0.95\textwidth]{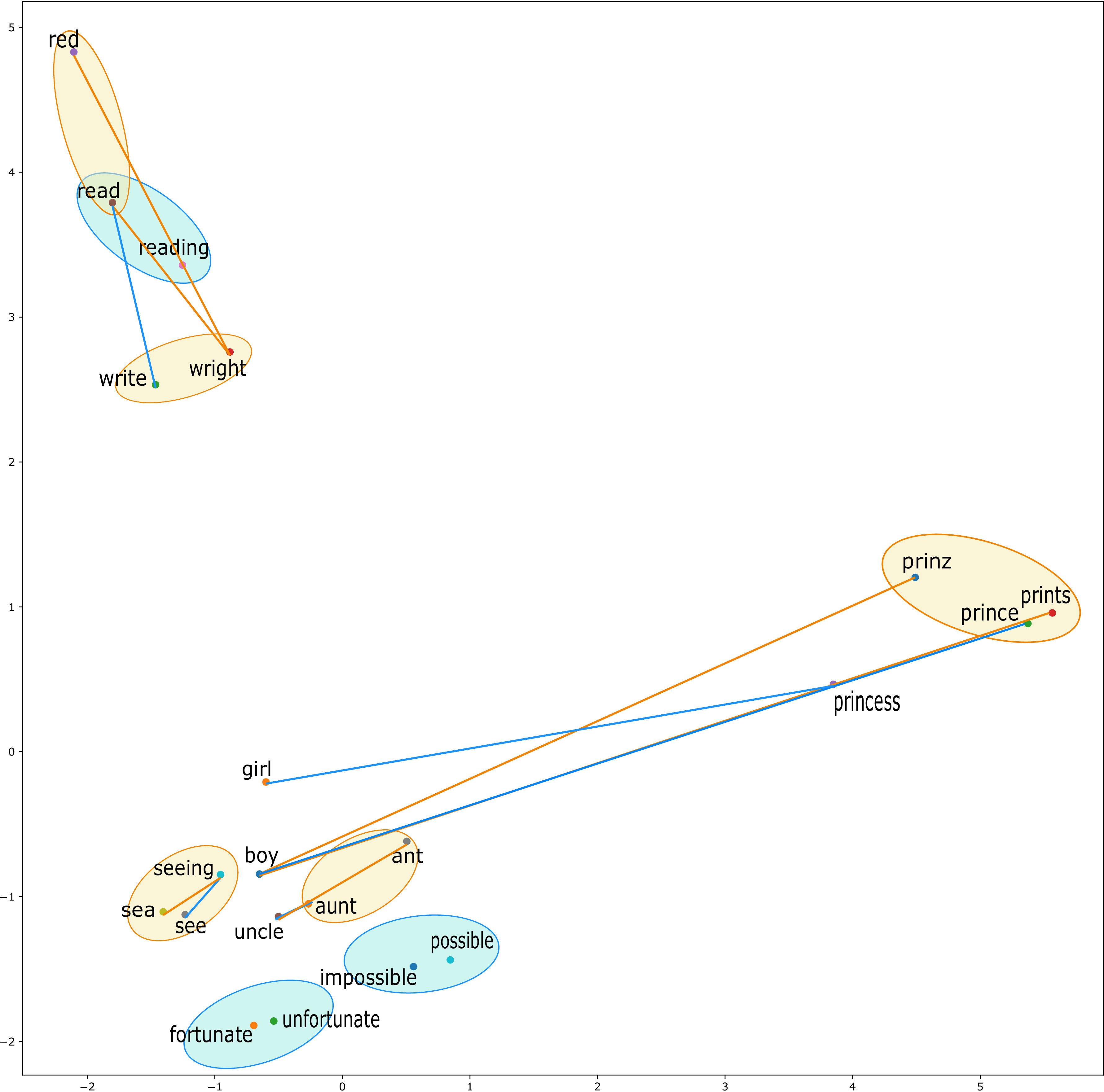}
\caption{Confusion2Vec 2.0: C2V-a}\label{fig:pca_c2v_a}
\end{subfigure}
\caption{2-D plots of selected word vectors portraying semantic, syntactic and acoustic relationships after dimension reduction using PCA\\
\small{The blue lines indicate semantic relationships, blue ellipses indicate syntactic relationships, red lines indicate acoustic-semantic/syntactic relations and red ellipses indicate acoustic ambiguity word relations.}}\label{fig:pca}
\end{figure*}

\subsection{Model Concatenation}
Further, the confusion2vec model can be concatenated with the other word embedding models to produce a new word vector space that can result in better representations as seen in \cite{shivakumar2019confusion2vec}.
Table~\ref{tab:concat_results} lists the results of the concatenated models.
For the previous, non-subword version of the confusion2vec, the vector models are concatenated with the word2vec model trained on the ASR output transcripts (C2V-1).
The choice of using the C2V-1 instead of the Google W2V for concatenation was based on empirical findings.
Where as to maintain compatibility of subword encoding, the confusion2vec 2.0 models are concatenated with fastText models.

First, comparisons between the non-concatenated versions in Table~\ref{tab:results} and the concatenated version in Table~\ref{tab:concat_results}, of the non-subword models, we observe an improvement of approximately 7.22\% relative in average analogy accuracy after concatenation.
We don't observe significant improvement with subword based models after concatenation in terms of average analogy accuracy.
However, we observe different dynamics between the acoustic ambiguity and the semantic and syntactic subspaces.
Concatenation results in improved semantic and syntactic evaluations with the expense of drop in accuracies of acoustic analogy task.
We also note improvements (9.27\% relative) in semantic\&syntactic-acoustic analogy task after concatenation confirming meaningful existence of both ambiguity and semantic-syntactic relations.
Moreover, the word similarity task also yields better correlation after concatenation. 

Next, comparisons of the confusion2vec 1.0 (non-subword) and the subword version, we observe significant improvements in the semantic\&symantic analogy task (7.51\% relative) as well as the semantic\&syntactic-acoustic analogy tasks (21.78\% relative).
Moreover, the subword models outperform the non-subword version in both of the similarity tasks.
The subword models slightly under-perform in the acoustic analogy task, but more crucially outperform the Google W2V and FastText baselines significantly.

Further, the concatenated models can be fine-tuned and optimized to exploit additional gains as found in \cite{shivakumar2019confusion2vec}.
The row corresponding to Confusion2Vec 1.0 - C2V + C2V-c (JT) is the best result obtained in \cite{shivakumar2019confusion2vec} which involves 2-passes.
The Confusion2Vec 2.0 with the subword modeling with a single pass training gives comparable performance to the 2-pass approach.
Thus we skip the 2-pass approach with the subword model in favor of ease of training and reproducibility.

\subsection{Embedding Visualization}
Figure~\ref{fig:pca} illustrates the word vector spaces of fastText embeddings and the proposed C2V-a embeddings after dimension reduction using principal component analysis.
We observe meaningful interactions between the semantic\&syntactic subspace and the acoustic ambiguity subspace.
For example, in Figure~\ref{fig:pca_c2v_a}, vectors \emph{``boy''-``prince''}, \emph{``see''-``seeing''}, \emph{``read''-``write''}, \emph{``uncle''-``aunt''} are similar to acoustically ambiguous vector \emph{``boy''-``prints''}, \emph{``sea''-``seeing''}, \emph{``read''-``write''}, \emph{``uncle''-``ant''} respectively which is not the case in Figure~\ref{fig:pca_fasttext} with fastText embeddings.
Such vector relationships can be exploited for downstream spoken language applications by providing crucial acoustic ambiguity information to recover from speech recognition errors.
Also note, the acoustically ambiguous words such as \emph{``prinz''}, \emph{``prince''}, \emph{``prints''} are found clustered together.
Another important observation is that the word \emph{``prinz''}, out-of-vocabulary in English, has an orphaned representation under fastText in Figure~\ref{fig:pca_fasttext}.
However, \emph{``prinz''} finds a meaningful representation on the basis of acoustic signature in the proposed Confusion2vec model as seen in Figure~\ref{fig:pca_c2v_a}, i.e., \emph{``prinz''} is clustered together with acoustically similar words \emph{``prince''} \& \emph{``prints''} and the vector \emph{``boy''-``prinz''} is similar to vector \emph{``boy''-``prince''}.
Occurrence of out-of-vocabulary words such as \emph{``prinz''} is common place with end-to-end ASR systems that output characters prone to errors.
Note, out-of-vocabulary words such as \emph{``prinz''} cannot be represented by typical word embeddings such as word2vec, GloVe etc and hence sub-optimal for representation with many end-to-end ASR systems.

\section{Spoken Language Intent Detection}\label{sec:slu}
In this section, we apply the proposed word vector embedding to the task of spoken language intent detection.
Spoken language intent detection is the process of decoding the speaker's intent in contexts involving voice commands, call routing and any human computer interactions.
Many spoken language technologies use an ASR to convert the speech signal to text, a process prone to errors such as due to the varying speaker and noise environments.
The erroneous ASR outputs in turn result in degradation of the downstream intent classification.
Few efforts have focused on handling the errors of the ASR to make the subsequent intent detection process more robust to errors.
These efforts often involve training the intent classification systems on noisy ASR transcripts.
The downsides of training the intent classifiers on the ASR is that the systems are limited with the amount of speech data available.
Moreover, varying speech signal conditions and use of different ASR models make such classifiers non-optimal and less practical.
In many scenarios, speech data is not available to enable adaptation on ASR transcripts.

In our previous work \cite{shivakumar2019spoken}, we applied the non-subword version of the Confusion2vec to the task of spoken language intent detection.
We demonstrated the Confusion2vec is able to perform as efficiently as the popular word embeddings like word2vec and GloVe on clean manual transcripts giving comparable classification error rates.
More importantly, we were able to illustrate the robustness of the confusion2vec embeddings when evaluated on the noisy ASR transcripts.
The confusion2vec gives significantly better accuracies (upto relative 20\% improvements) when evaluated on ASR transcripts compared to the word2vec, GloVe embeddings and state-of-the-art models involving more complex neural network intent classification architectures.
Moreover, we also showed that Confusion2vec suffers the least degradation between clean and ASR transcripts.
We also found that the Confusion2vec consistently provides the best classification rates even when the intent classifier is trained on ASR transcripts.
The experiments indicated that the loss in accuracies between training the intent classifier on clean versus the ASR transcripts is reduced to 0.89\% from 2.57\% absolute.
Overall, the results illustrate that confusion2vec has inherent knowledge of the acoustic ambiguity (similarity) word relations which correlates with the ASR errors using which the classifier is able to recover from certain errors more efficiently. 
In this section, we incorporate the confusion2vec 2.0 embeddings with inherent knowledge of acoustic ambiguity to allow robust intent classification.

\subsection{Database}
We conduct experiments on the Airline Travel Information Systems (ATIS) benchmark dataset \cite{hemphill1990atis}.
The dataset consists of humans making flight related inquiries with an automated answering machine with audio recorded and its transcripts manually annotated.
ATIS consists of 18 intent categories.
The dataset is divided into train (4478 samples), development (500 samples) and test (893 samples) consistent with previous works \cite{shivakumar2019spoken,hakkani2016multi, goo2018slot}.
For ASR evaluations, the audio recordings are down-sampled from 16kHz to 8kHz and then decoded using the ASR setup described in section~\ref{sec:asr} using the audio mappings\footnote{\texttt{https://github.com/pgurunath/slu\_confusion2vec}}.
The ASR achieves a WER of 18.54\% on the ATIS test set.

\subsection{Experimental Setup}
For intent classification we adopt a simple RNN architecture identical to \cite{shivakumar2019spoken}, to allow for direct comparison.
The architecture of the neural network is intentionally kept simple for effective inference of the efficacy of the proposed embedding word features.
The classifier is comprised of an embedding layer followed by a single layer of bi-directional recurrent neural network (RNN) with long short-term memory (LSTM) units which is followed by a linear dense layer with softmax function to output a probability distribution across all the intent categories.
The embedding layer is fixed throughout the training except for the randomly initialized embeddings where the embedding is estimated on the in-domain data specific to the task of intent detection.

The intent classification models are trained on the 4478 samples of training subset and the hyper-parameters are tuned on the development set.
We choose the best set of hyper-parameters yielding the best results on the development set and then apply it on the unseen held-out test subset of both the manual clean transcripts and the ASR transcripts and report the results.
For training we treat each utterance as a single sample (batch size = 1).
The hyper-parameter space we experiment are as follows: the hidden dimension size of the LSTM is tuned over $\{32, 64, 128, 256\}$, the learning rate over $\{0.0005, 0.001\}$, the dropout is tuned over $\{0.1, 0.15, 0.2, 0.25\}$.
The Adam optimizer is employed for optimization and trained for a total of 50 epochs with early stopping when the loss on the development set doesn't improve for 5 consecutive epochs.

\subsection{Baselines} 
We include results from several baseline systems for providing comparisons of Confusion2Vec 2.0 with the popular context-free word embeddings, contextual embeddings, popular established NLU systems and the current state-of-the-art.
\begin{enumerate}
\item \textbf{Context-Free Embeddings}: GloVe\footnote{\texttt{https://nlp.stanford.edu/projects/glove/}} \cite{pennington2014glove}, skip-gram word2vec\footnote{\texttt{https://code.google.com/archive/p/word2vec/}} \cite{mikolov2013distributed} and fastText\footnote{\texttt{https://fasttext.cc/docs/en/pretrained-vectors.html}} \cite{bojanowski2017enriching} word representations are employed.
They are referred to as context-free embeddings since the word representations are static irrespective of the context.

\item \textbf{ELMo}: \citet{peters2018deep} proposed deep contextualized word representation based on character based deep bidirectional language model trained on large text corpus.
The models effectively model syntax and semantics of the language along varying linguistic contexts.
Unlike context-free embeddings, ELMo embeddings have varying representations for each word depending on the word's context.
We employ the original model trained on 1 Billion Word Benchmark with 93.6 million parameters\footnote{\texttt{https://allennlp.org/elmo}}.
For intent-classification we add a single bi-directional LSTM layer with attention for multi-task joint intent and slot predictions.
\item \textbf{BERT}: \citet{devlin2018bert} introduced BERT bidirectional contextual word representations based on self attention mechanism of Transformer models.
BERT models make use of masked language modeling and next sentence prediction to model language.
Similar to ELMo, the word embeddings are contextual, i.e., vary according to the context.
We employ ``bert-base-uncased'' model\footnote{\texttt{https://github.com/google-research/bert}} with 12 layers of 768 dimensions each trained on BookCorpus and English Wikipedia corpus.
For intent-classification we add a single bi-directional LSTM layer  with attention for multi-task joint intent and slot predictions.
\item \textbf{Joint SLU-LM}: \citet{liu2016joint} employed joint modeling of the next word prediction along with intent and slot labeling.
The unidirectional RNN model updates intent states for each word input and uses it as context for slot labeling and language modeling.
\item \textbf{Attn. RNN Joint SLU}: \citet{Liu+2016} proposed attention based encoder-decoder bidirectional RNN model in a multi-task model for joint intent and slot-filling tasks.
A weighted average of the encoder bidirectional LSTM hidden states provides information from parts of the input word sequence which is used together with time aligned encoder hidden state for the decoder to predict the slot labels and intent.
\item \textbf{Slot-Gated Attn.}: \citet{goo2018slot} introduced a slot-gated mechanism which introduces additional gate to improve slot and intent prediction performance by leveraging intent context vector for slot filling task.
\item \textbf{Self Attn. SLU}: \citet{li2018self} proposed self-attention model with gate mechanism for joint learning of intent classification and slot filling by utilizing the semantic correlation between slots and intents.
The model estimates embeddings augmented with intent information using self attention mechanism which is utilized as a gate for slot filling task.
\item \textbf{Joint BERT}: \citet{chen2019bert} proposed to use BERT embeddings for joint modeling of intent and slot-filling.
The pre-trained BERT embeddings are fine-tuned for (i) sentence prediction task - intent detection, and (ii) sequence prediction task - slot filling.
The Joint BERT model lacks the bi-directional LSTM layer in comparison to the earlier baseline \emph{BERT} based model.
\item \textbf{SF-ID Network}: \citet{e2019novel} introduced a bi-directional interrelated model for joint modeling of intent detection and slot-filling.
An iteration mechanism is proposed where the SF subnet introduces the intent information to slot-filling task while the ID-subnet applies the slot information to intent detection task.
For the task of slot-filling a conditional random field layer is used to derive the final output.
\item \textbf{ASR Robust ELMo}: \citet{huang2020learning} proposed ASR robust contextualized embeddings for intent detection.
ELMo embeddings are fine-tuned with a novel loss function which minimizes the cosine distance between the acoustically confused words found in ASR confusion networks.
Two techniques based on supervised and unsupervised extraction of word confusions are explored.
The fine-tuned contextualized embeddings are then utilized for spoken language intent detection.
\end{enumerate}

\begin{table*}[t]
  \centering
  \begin{tabular}{llccc}
    \toprule
    & \textbf{Model}   & \textbf{Reference}    & \textbf{ASR}  & \textbf{$\Delta_{\text{diff}}$} \\
    \midrule
    \multirow{10}{*}{\shortstack[l]{Context-Free Embeddings}} & Random	& 2.69	& 10.75	& 8.06 \\
    & GloVe \cite{pennington2014glove}        & 1.90          & 8.17          & 6.27 \\
    & Word2Vec \cite{mikolov2013distributed}  & 2.69          & 8.06          & 5.37 \\
		& fastText \cite{bojanowski2017enriching} & 1.90					& 8.40					& 6.50 \\
		& Joint SLU-LM \cite{liu2016joint}~$^{\dagger}$        & 1.90          & 9.41          & 7.51 \\
    & Attn. RNN Joint SLU \cite{Liu+2016}~$^{\dagger}$     & 1.79  				& 8.06          & 6.27 \\
    & Slot-Gated Attn. \cite{goo2018slot}~$^{\dagger}$     & 3.92          & 10.64         & 6.72 \\
    & Self Attn. SLU \cite{li2018self}~$^{\dagger}$        & 2.02          & 9.18          & 7.16 \\
		& SF-ID Network \cite{e2019novel}~$^{\dagger}$					& 3.14					& 10.53					& 7.39 \\
		& C2V 1.0 \cite{shivakumar2019confusion2vec} & 2.46       & 6.38          & 3.92 \\
		\midrule
		\multirow{5}{*}{\shortstack[l]{Contextual Embeddings}} & ELMo \cite{peters2018deep}~$^{\dagger}$	& 1.46	& 7.05	& 5.59 \\
		& BERT \cite{devlin2018bert}~$^{\dagger}$						& \textbf{1.12}	& 6.16					& 5.04 \\
		& Joint BERT \cite{chen2019bert}~$^{\dagger}$					& 2.46					& 7.73					& 5.27 \\
		& ASR Robust ELMo (unsup.) \cite{huang2020learning} & 3.24		& 5.26	& 2.02 \\
		& ASR Robust ELMo (sup.) \cite{huang2020learning} & 3.46		& 5.03	& \textbf{1.57} \\
		\midrule
		\multirow{4}{*}{\shortstack[l]{Proposed Context-Free Embeddings}} & C2V-c 2.0	& 3.36	& 5.82	& 2.46 \\
		& C2V-a 2.0																& 2.46					&\textbf{4.37}	& \textbf{1.91} \\
		& fastText + C2V-c 2.0										& 1.79					&\textbf{4.70} 	& 2.91 \\
		& fastText + C2V-a 2.0										& 1.90					& 5.04 & 3.14 \\
    \midrule
    \bottomrule
  \end{tabular}
	\captionsetup{justification=justified}
  \caption[]{Results: Model trained on clean Reference: Classification Error Rates (CER) for Reference and ASR Transcripts\\
	\small{$\Delta_{\text{diff}}$ is the absolute degradation of model from clean to ASR. C2V 1.0 corresponds to C2V-1 + C2V-c (JT) in Table~\ref{tab:results} and~\ref{tab:concat_results}.\\$\dagger$ indicates joint modeling of intent and slot-filling.}}\label{tab:slu_results}
\end{table*}

\subsection{Results}
In this section, we conduct experiments by training models on (i) clean human annotations and (ii) noisy ASR transcriptions.
\subsubsection{Training on Clean Transcripts}
Table~\ref{tab:slu_results} lists the results of the intent detection in terms of classification error rates (CER).
The ``Reference'' column corresponds to results on human transcribed ATIS audio and the ``ASR'' corresponds to the evaluations on the noisy speech recognition transcripts.
Firstly, evaluating on the Reference clean transcripts, we observe the confusion2vec 2.0 with subword encoding is able to achieve the third best performance.
The best performing confusion2vec 2.0 achieves a CER of 1.79\%.
Among the different versions of the proposed subword based confusion2vec, we find that the concatenated versions are slightly better.
We believe this is because the concatenated models exhibit better semantic and syntactic relations (see Table~\ref{tab:results} and ~\ref{tab:concat_results}) compared to the non-concatenated ones.
Among the baseline models, the contextual embedding like BERT and ELMo gives the best CER.
Note, the proposed confusion2vec embeddings are context-free and are able to outperform other context-free embedding models such as GloVe, word2vec and fastText.

Secondly, evaluating the performance on the noisy ASR transcripts, we find that all the subword based confusion2vec 2.0 models outperform the popular word vector embeddings by a big margin.
The subword-confusion2vec gives an improvement of approximately 45.78\% relative to the best performing context-free word embeddings.
The proposed embeddings also improve over the contextual embeddings including BERT and ELMo (relative improvements of 29.06\%).
Moreover, the results are also an improvement over the non-subword confusion2vec word vectors (31.50\% improvement).
Comparisons between the different versions of the proposed confusion2vec show the intra-confusion configuration yields the least CER.
The best results with the proposed model outperforms the state-of-the-art (ASR Robust ELMo \cite{huang2020learning}) by reducing the CER by a relative of 13.12\%.
Inspecting the degradation, $\Delta_{\text{diff}}$ (drop in performance between the clean and ASR evaluations), we find that all the confusion2vec 2.0 with subword information undergo low degradation while giving the best CER, thereby re-affirming the robustness to noise in transcripts.
This confirms our initial hypothesis that the subword encoding is better able to represent the acoustic ambiguities in the human language.

\subsubsection{Training on Noisy ASR Transcripts}
Table~\ref{tab:slu_asr_results} presents the results obtained by training models on the ASR transcripts and evaluated on the ASR transcripts.
Here we omit all the joint intent-slot filling baseline models, since training on ASR transcripts need aligned set of slot labels due to insertion, substitution and deletion errors which is out-of-scope of this study.
We note that the confusion2vec models give significantly lower CER.
The subword based confusion2vec models also provide improvements over the non-subword based confusion2vec model (21.28\% improvement).
Comparing the results in Table~\ref{tab:slu_results} and Table~\ref{tab:slu_asr_results}, we would like to highlight the subword-confusion2vec model gives a minimum CER of 4.37\% on model trained on clean transcripts which is much better than the CER obtained by popular word embeddings like word2vec, GloVe, fastText even when trained on the ASR transcripts (15.15\% better relatively).
These results prove the subword-confusion2vec models can eliminate the need for re-training natural language understanding and processing algorithms on ASR transcripts for robust performance.

\begin{table}[t]
  \centering
  \begin{tabular}{lll}
    \toprule
     \textbf{Model}   & \textbf{WER \%} &\textbf{CER \%}  \\
    \midrule
    Random                   								& 18.54   & 5.15 \\
    GloVe \cite{pennington2014glove}        & 18.54   & 6.94 \\
    Word2Vec \cite{mikolov2013distributed}  & 18.54   & 5.49 \\
    \citet{schumann2018incorporating}       & 10.55   & 5.04\footnotemark  \\
		
		C2V 1.0                					& 18.54   & 4.70 \\
		\midrule
		C2V-c 2.0																& 18.54					& 4.82 \\
		C2V-a 2.0																& 18.54					& \textbf{4.26} \\
		fastText + C2V-c 2.0										& 18.54 				& \textbf{3.70} \\
		fastText + C2V-a 2.0										& 18.54					& \textbf{4.26} \\
    \midrule
    \bottomrule
  \end{tabular}
	\captionsetup{justification=centering}
  \caption{Results: Model trained and evaluated on ASR transcripts.\\
	\small{C2V 1.0 corresponds to C2V-1 + C2V-c (JT) in Table~\ref{tab:results} and~\ref{tab:concat_results}}}\label{tab:slu_asr_results}
\end{table}
\footnotetext[11]{We don't domain-constrain, optimize or re-score our ASR, as in \cite{schumann2018incorporating}}

\section{Conclusion}\label{sec:conclusion}
In this paper, we proposed the use of subword encoding for modeling the acoustic ambiguity information and augment word vector representations along with the semantic and syntax of the language.
Each word in the language is represented as a sum of its constituent character n-gram subwords.
The advantages of the subwords are confirmed by evaluating the proposed models on various word analogy tasks and word similarity tasks designed to assess the effective acoustic ambiguity, semantic and syntactic knowledge inherent in the models.
Finally, the proposed subword models are applied to the task of spoken language intent detection.
The results of intent classification system suggest the proposed subword confusion2vec models greatly enhance the classification performance when evaluated on the noisy ASR transcripts.
The results highlight that subword-confusion2vec models are robust and domain-independent and do not need re-training of the classifier on ASR transcript.

In the future, we plan to model ambiguity information using deep contextual modeling techniques such as BERT.
We believe bidirectional information modeling with attention can further enhance ambiguity modeling.
On the application side, we plan to implement and assess the effect of using Confusion2vec models for a wide range of natural language understanding and processing applications such as speech translation, dialogue tracking etc.

\bibliographystyle{plainnat}
\bibliography{refs}

\end{document}